\documentclass[sigconf, nonacm]{acmart}

\usepackage{pvldb}

\renewcommand\vldbdoi{XX.XX/XXX.XX}
\renewcommand\vldbpages{XXX-XXX}
\renewcommand\vldbavailabilityurl{https://github.com/chris265100/TSSDA}

\newcommand\modelName{TSSDA}
\newcommand\BenchName{TERN}
\usepackage{tikz}
\usepackage{enumitem}

\usepackage[ruled,vlined,linesnumbered]{algorithm2e}

\usepackage{amsmath,bm}

\usepackage{multirow}
\usepackage{multicol}
\usepackage{booktabs}
\usepackage{tabularx}
\usepackage{array}
\usepackage{diagbox}
\usepackage{makecell}

\usepackage{graphicx}
\usepackage{svg}
\usepackage{subcaption}
\usepackage{pgfplots}
\usepackage{tkz-euclide}

\usepackage{stfloats}
\usepackage{verbatim}
\usepackage{marginnote}
\usepackage{tcolorbox}
\tcbuselibrary{breakable}
\usepackage{xcolor}
\usepackage{pifont}

\usepackage{todonotes}

\newcommand{\cmark}{\ding{51}}
\newcommand{\xmark}{\ding{55}}

\newcolumntype{C}{>{\centering\arraybackslash}m{1.5cm}}
\newcolumntype{Y}{>{\centering\arraybackslash}m{1.4cm}}
\begin{document}

\title{TSSDA: An Autonomous Agent for Constructing a Time-Series Text-to-SQL Benchmark}

\author{Zhihan Zheng}
\affiliation{%
  \institution{Tsinghua University, China}
  \country{}
}
\email{zhengzh26@mails.tsinghua.edu.cn}

\author{Shaoxu Song}
\affiliation{%
  \institution{Tsinghua University, China}
  \country{}
}
\email{sxsong@tsinghua.edu.cn}

\author{Haitao Yuan}
\affiliation{%
  \institution{NTU, Singapore}
  \country{}
}
\email{haitao.yuan@ntu.edu.sg}

\author{Minxiao Chen}
\affiliation{%
  \institution{BUPT, China}
  \country{}
}
\email{chenminxiao@bupt.edu.cn}








\begin{abstract}

As agents increasingly undertake data querying, Text-to-SQL serves as a foundational capability for database access. 
However, for text-to-SQL in time-series databases, high-quality benchmarks and large-scale training corpora remain scarce, limiting the systematic evaluation and domain adaptation of existing models and agents.
Existing automated SQL generation approaches lack sufficient autonomy, leaving key challenges unresolved. First, numerous combinations of temporal syntax and functions complicate comprehensive coverage of valid queries. Second, as the corpus grows, generating valuable and diverse new data becomes increasingly challenging. Third, schema and data heterogeneity hinder reuse of temporal query structures, limiting generation efficiency.
To address these challenges, we propose \textbf{\modelName{}}, a \textbf{T}ime-\textbf{S}eries \textbf{S}QL (TSSQL) Generation \textbf{D}ata \textbf{A}gent framework, and construct \textbf{\BenchName{}}, a comprehensive \textbf{T}ime-s\textbf{ER}ies text-to-SQL be\textbf{N}chmark.
First, to achieve broad coverage of valid queries, we systematically model TSSQL syntax and semantics to construct a data generation space. Next, to explore the space and improve diversity, we develop a Data Agent that autonomously orchestrates data generation. 
The agent comprises a planning and decision-making module and a query generation module, with online adaptation for continuous self-improvement. Finally, we introduce a temporal-aware memory that enables adaptive reuse of experience across heterogeneous data instances, improving generation efficiency. 
Comprehensive evaluations on \BenchName{} reveal the limitations of existing methods in performing Text-to-TSSQL tasks. 
Furthermore, extensive experiments demonstrate that \modelName{} substantially improves coverage, diversity, and complexity.

\end{abstract}

\maketitle

\vldbtopmatter

\section{Introduction}

\begin{figure}
  \centering  
  \includegraphics[width=0.99\linewidth]{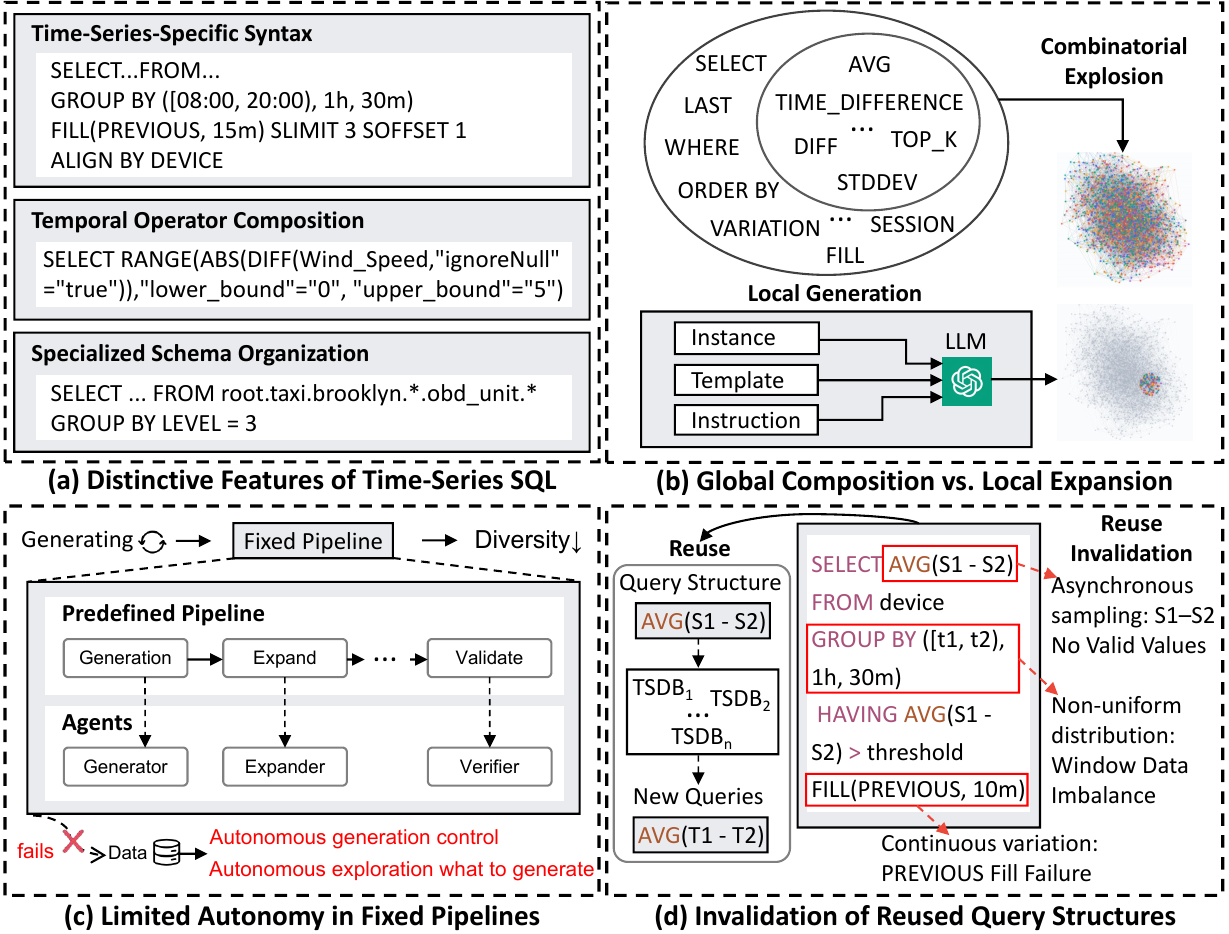}
  \vspace{-0.1in}
  \caption{Challenges of Text-to-SQL for TSDBs}
  \label{fig:intro}
\end{figure}

With advances in large language models (LLMs), agents increasingly undertake data querying~\cite{TiInsight, APEX-SQL} and processing~\cite{AutoPrep, SiriusBI} tasks. Text-to-SQL, which translates natural language requests into SQL queries, is becoming a foundational capability for these agents~\cite{NL2SQLSurvey, NLSQLSurvey2}. Numerous approaches~\cite{OpenSearch-SQL, OmniSQL, DeepEye-SQL} have emerged, alongside increasingly challenging benchmarks~\cite{VLD-Bench, TACO, Spider2}. However, existing studies primarily focus on relational databases, leaving Text-to-SQL for time-series databases (TSDBs) largely underexplored.

SQL in TSDBs differs fundamentally from conventional SQL in both query semantics and language constructs. As shown in Fig.~\ref{fig:intro}(a), we highlight several distinctive features of time-series SQL (TSSQL). First, TSDBs support diverse time-series-specific syntax, including temporal windows, missing-value filling, and result alignment, as illustrated in the example. Second, TSSQL often composes multiple functions and operators for complex temporal analytics. The example detects absolute wind-speed changes exceeding 5 between consecutive valid readings, while real-world queries may be more complex. Third, TSDBs exhibit diverse schema organizations and are tightly coupled with time-series syntax. For example, we illustrate IoTDB's schema~\cite{IoTDB} with a query aggregating matching OBD sensor readings by vehicle in Brooklyn. Beyond SQL structure, time-series queries also embody rich and distinctive temporal semantics. Consequently, existing Text-to-SQL methods designed for relational databases are inapplicable to TSDBs. 

To comprehensively evaluate Text-to-SQL for TSDBs and provide a domain-specific training corpus, there is a pressing need for a high-quality, large-scale benchmark. Although recent efforts~\cite{TS-NL2SQL, TQTS} have made initial attempts, they often rely on manual curation or direct prompting, limiting data coverage, diversity, and scalability. Driven by recent developments in agentic AI, data agents~\cite{DAgentSurvey} have emerged as a promising paradigm for autonomously handling complex data tasks. However, existing data agents primarily focus on managing~\cite{AgentTune}, preparing~\cite{AutoPrep, DeepPrep}, and analyzing~\cite{SiriusBI, ST-Raptor} existing data, with limited attention to the autonomous generation of domain-specific data, such as NL–TSSQL pairs. Unlike general-purpose data generation, TSSQL data generation is subject to strict syntactic and semantic constraints, demanding higher data quality. Meanwhile, existing LLM- and agent-based Text-to-SQL data generation approaches~\cite{VLD-Bench,TACO,OmniSQL,SQL-Factory} lack sufficient autonomy for generating NL–TSSQL pairs, leaving several key challenges unresolved:

\textbf{Challenge 1: Modeling Valid Compositions in TSSQL.} TSSQL introduces a variety of specialized temporal syntax and numerous time-series functions. As shown in Fig.~\ref{fig:intro}(b), directly taking the Cartesian product leads to a combinatorial explosion, while only a small subset is syntactically valid, semantically meaningful, and compatible with the schema instance. Existing manually constructed datasets~\cite{BIRD, Spider2, TQTS} require experts or annotators, which are costly and difficult to scale, while automated approaches primarily rely on instance transformation~\cite{VLD-Bench, CLARITY}, template instantiation~\cite{ScienceBenchmark, TACO}, or instruction-guided synthesis~\cite{TS-NL2SQL, OmniSQL, SQL-Factory}. Instance- and template-based methods constrain generation to predefined structural neighborhoods, whereas instruction-guided methods are highly sensitive to the quality of manually specified task requirements and constraints, leaving query validity largely dependent on LLM reasoning and post-hoc verification. Consequently, existing approaches lack systematic modeling of valid SQL compositions, limiting both coverage and validity.

\textbf{Challenge 2: Autonomous Orchestration for TSSQL Diversity.} Data generation is inherently dynamic: as the corpus evolves, the utility of new data changes with the knowledge and characteristics represented in the generated data. A key challenge is therefore to determine what is most valuable to explore among numerous valid candidates at each generation step to enhance data diversity. As shown in Fig.~\ref{fig:intro}(c), existing agent-based approaches~\cite{SQL-Factory, SemanticAgent, ABISS} introduce multi-agent collaboration and environmental interaction, yet their agents are typically embedded in predefined generation pipelines, with autonomy largely confined to executing or coarsely selecting among predefined procedures such as generation, expansion, and validation. Such coarse-grained autonomy lacks fine-grained control over the characteristics of generated data and cannot dynamically assess the marginal utility of potential data additions based on the evolving corpus. Consequently, these methods may repeatedly revisit well-covered regions while under-exploring more informative combinations, limiting the diversity and complexity of generated data.


\textbf{Challenge 3: Efficiently Reusing Temporal Semantics.} TSSQL generation instantiates recurring local temporal semantics across diverse database entities, creating a pronounced structure–instance asymmetry: similar semantic structures must be repeatedly reconstructed across heterogeneous data instances. However, the applicability of TSSQL structures depends not only on schema compatibility but also on data characteristics, such as sampling granularity, temporal distribution, and synchronization. As illustrated in Fig.~\ref{fig:intro}(d), asynchronous sampling invalidates point-wise operations, non-uniform distribution undermines time-window aggregation, and continuous variation requires linear rather than previous filling. Existing approaches~\cite{RingSQL, SQLForge} typically achieve reuse by extracting query-level structural SQL templates. For TSSQL, however, structural compatibility does not imply temporal applicability, while coarse SQL-level reuse fails to capture local temporal structures.

To comprehensively address the aforementioned challenges, we propose a novel data generation framework, \textbf{\modelName{}}, and construct \textbf{\BenchName{}}, a large-scale Text-to-TSSQL benchmark. \textbf{To address the first challenge}, we systematically model TSSQL syntax and semantics to construct a data generation space. We decompose TSSQL into core syntactic modules with explicit semantic decision variables, and model their cross-module dependencies. The resulting space integrates schema information, dependency propagation, and evolving semantic states to provide valid decision paths for composing diverse TSSQL queries. \textbf{To address the second challenge}, we develop a \textit{Data Agent} that autonomously navigates, explores, and orchestrates the data space. Specifically, we design dozens of Skills that provide the Agent with comprehensive autonomous capabilities. The Data Agent consists of two collaborative agents: a \textit{Semantic Decision Agent} that dynamically plans Skill execution and makes semantic decisions, and a \textit{Query Generation Agent} that incrementally materializes the temporal query. An online adaptation mechanism further enables the Agent to adapt to the evolving data distribution and accumulated experience, continuously improving generation quality and diversity. \textbf{To address the third challenge}, we propose a temporal-aware hierarchical memory for efficient reuse of generation experience. We abstract data characteristics and generation states into temporal profiles as memory indices. A hierarchical memory then retrieves historical semantic decisions and operator expressions through temporal profile matching and reliability-aware selection. An adaptive management mechanism further dynamically updates the accumulated experience.

In summary, we make the following contributions:
\vspace{-\topsep}
\begin{itemize}[leftmargin=10.2pt]
\setlength{\itemsep}{0pt}
\setlength{\parsep}{0pt}
\setlength{\parskip}{0pt}
\item We formalize the problem of automated Text-to-SQL data generation for TSDBs and propose a highly autonomous Data Agent framework for constructing high-quality data. (Sec.~\ref{sec:2} and ~\ref{sec:3}).
\item We extract core syntax modules from TSSQL and model their semantic dependencies to construct an executable and explorable data generation space. (Sec.~\ref{sec:4}).
\item We propose a Data Agent with carefully designed skills, exploration algorithms, and online adaptation mechanisms to autonomously explore and orchestrate the data space. (Sec.~\ref{sec:5})
\item We develop a temporal-aware memory with temporal pattern profiling and adaptive memory management to efficiently reuse local temporal semantics. (Sec.~\ref{sec:6}).
\item We construct \textbf{\BenchName{}}, a large-scale benchmark built on 32 real-world industrial datasets and comprising 13,182 unique NL–TSSQL pairs. We conduct extensive evaluations on \textbf{\BenchName{}} and comprehensively compare \textbf{\modelName{}} against baseline methods. (Sec.~\ref{sec:7})
\end{itemize}

\section{Preliminaries}
\label{sec:2}

\subsection{Problem Formulation}
\noindent\textbf{Basic Definitions.}
We first introduce the basic concepts related to time-series databases (TSDBs).

\begin{itemize}[leftmargin=10.2pt]
\setlength{\itemsep}{0pt}
\setlength{\parsep}{0pt}
\setlength{\parskip}{0pt}

\item \textit{Time Series}: A time series is an ordered sequence of observations, each consisting of a timestamp and its corresponding value, where time serves as the primary dimension in TSDBs.

\item \textit{Schema}: A TSDB schema $\mathcal{S}$ describes the organization of time-series data under a specific data model, such as tree-based~\cite{IoTDB} or tag-based~\cite{influxDB} models, including the associated entities, devices, measurements, and their structural relationships.

\item \textit{Temporal Operators}:Temporal operators \(\mathcal{O}_T\) are query operators or functions applied to time series for various temporal analysis tasks. They mainly include time-series generating functions (TSGFs) for sequence transformation and aggregate functions (AGGs) for sequence aggregation.

\item \textit{TSSQL}: TSSQL refers to SQL-like query languages provided by TSDBs for querying and analyzing time-series data. A TSSQL query $q$ specifies target time series, temporal semantics, and optionally temporal operators from $\mathcal{O}_T$. Its execution under schema $\mathcal{S}$ is denoted as
$\operatorname{Exec}(q,\mathcal{S})\rightarrow Y$,
where $Y$ is the result.

\end{itemize}

\noindent \textbf{NL2TSSQL.} Given a natural language question $u$ and a time-series schema $\mathcal{S}$, the NL2TSSQL task aims to translate $u$ into a TSSQL query $q$ that correctly expresses the intended query semantics. Formally, the task can be expressed as $f_{\mathrm{NL2TSSQL}}:(u,\mathcal{S},\mathcal{O}_T,\Gamma_T)\rightarrow q,$
where $\Gamma_T$ denotes the temporal grammar and syntactic constraints imposed by the target TSSQL dialect. The generated query $q$ should conform to schema $\mathcal{S}$, the supported temporal operator space $\mathcal{O}_T$, and the temporal grammar constraints $\Gamma_T$, while faithfully preserving the semantics specified by the natural language question $u$.

\noindent \textbf{Data Generation.}
Given the schema $\mathcal{S}$ of a target TSDB instance, the data generation task aims to generate a set of NL2TSSQL instances $\mathcal{Z}=\{(u_i,q_i)\}_{i=1}^{N}$, where $u_i$ is a natural language question and $q_i$ is its corresponding TSSQL query. Formally, the task is expressed as
$f_{\mathrm{DG}}:(\mathcal{S},\mathcal{O}_T,\Gamma_T)\rightarrow\mathcal{Z}$.
Each generated pair $(u_i,q_i)$ should be semantically consistent, while $q_i$ should conform to $\mathcal{S}$, $\mathcal{O}_T$, and $\Gamma_T$, and be executable on the target TSDB.

\subsection{Benchmark Definition}
\noindent \textbf{\BenchName} \textbf{Benchmark.}
\BenchName{} is an NL2TSSQL benchmark defined as $\mathcal{B}=(\mathbb{E},\mathcal{Z},\mathcal{R},\Pi)$, consisting of the following components:
\begin{itemize}[leftmargin=10.2pt]
\setlength{\itemsep}{0pt}
\setlength{\parsep}{0pt}
\setlength{\parskip}{0pt}

\item \textit{TSDB Environments}: $\mathbb{E}=\{(\mathcal{S}_j,\mathcal{O}_{T,j},\Gamma_{T,j})\}_{j=1}^{N}$,
representing the TSDB environments, where $N$ is the number of TSDB instances.

\item \textit{NL2TSSQL Instances}: $\mathcal{Z}=\bigcup_{j=1}^{N}\mathcal{Z}_j$, where 
$\mathcal{Z}_j=\{(u_{j,i},q_{j,i})\}_{i=1}^{n_j}$ denotes the set of NL--TSSQL pairs associated with environment $j$, and each pair $(u_{j,i},q_{j,i})$ satisfies the validity requirements.

\item \textit{Generation Specification}: 
$\mathcal{R}$ is an optional set of requirements that specifies the desired benchmark construction targets and enables data generation under customized objectives.

\item \textit{Evaluation Protocol}: 
$\Pi$ specifies the data partitioning strategy and evaluation settings of the benchmark.

\end{itemize}

Beyond instance-level validity, a benchmark should additionally satisfy dataset-level requirements on \textit{coverage}, \textit{diversity}, and \textit{quality} to enable comprehensive and reliable evaluation of NL2TSSQL systems.

\section{Framework Overview}
\label{sec:3}

\begin{figure*}
  \centering  
  \includegraphics[width=0.96\linewidth]{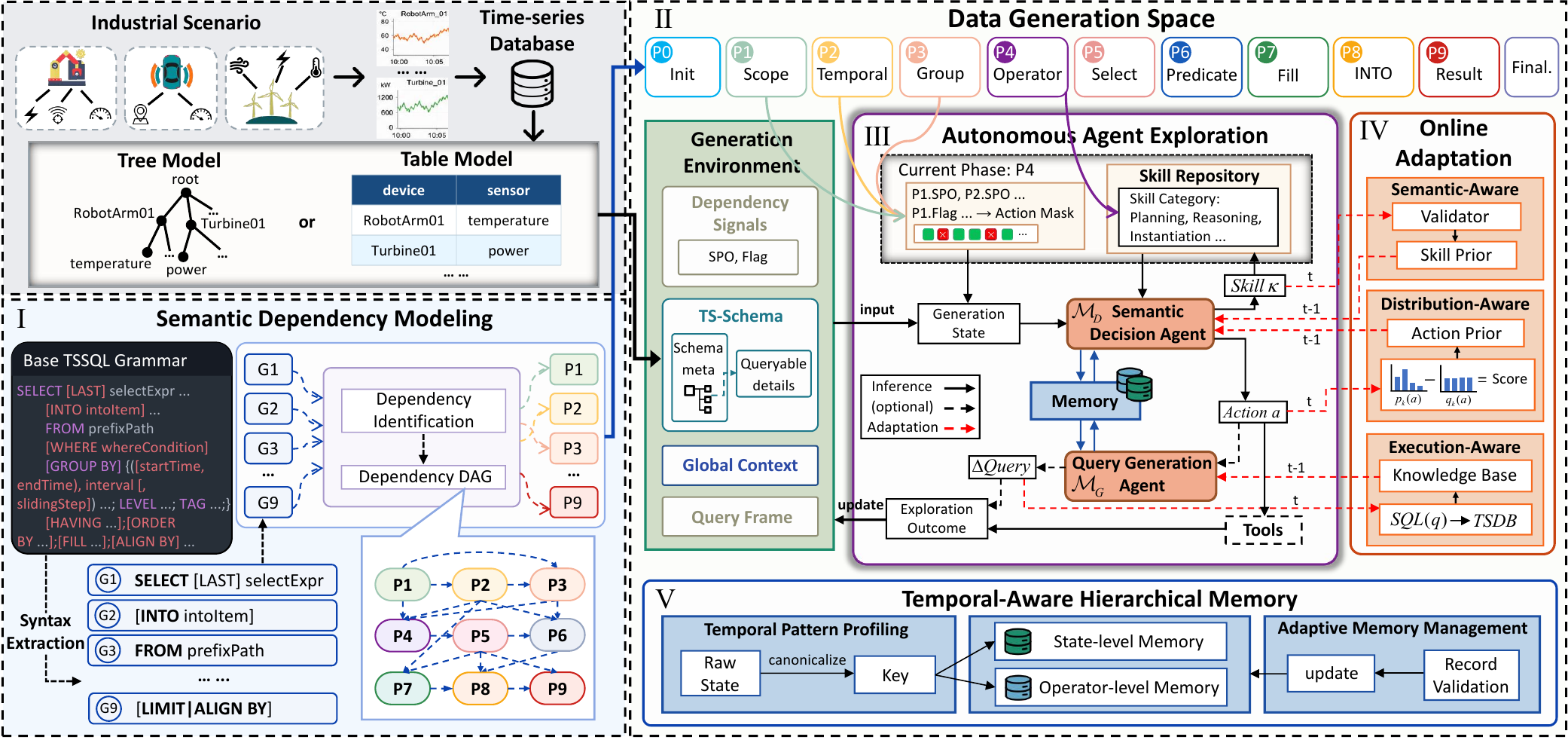}
  \vspace{-0.1in}
  \caption{Architecture of the \textbf{\modelName} Framework}
  \label{fig:framework}
\end{figure*}

In this section, we present the design of our framework, \textbf{\modelName}. Unlike existing automated NL2SQL data generation approaches, which primarily rely on instance transformation~\cite{VLD-Bench, CLARITY}, template instantiation~\cite{ScienceBenchmark, TACO}, or predefined workflows~\cite{OmniSQL, SQL-Factory, ABISS}, our method constructs an interactive data generation space by explicitly modeling syntax and semantic dependencies, and develops a Data Agent system that autonomously explores, orchestrates, and adapts this data infrastructure. As illustrated in Figure~\ref{fig:framework}, we collect diverse industrial time-series datasets as TSDB instances. \textbf{\modelName} comprises five key components, detailed below.  

\noindent \underline{\textbf{I. Semantic Dependency Modeling.}} Automated TSSQL generation should jointly ensure broad compositional coverage and strong validity. Existing approaches~\cite{TACO, SQL-Factory, ABISS} largely rely on prompt-specified rules and post-hoc augmentation, leaving the definition of valid query compositions implicit and requiring costly filtering to remove invalid queries. To address this issue, we model the TSSQL generation process at a fine granularity from both syntactic and semantic perspectives. Specifically, we first extract nine core syntactic modules from the base TSSQL grammar. We then analyze temporal dependencies and constraints from temporal semantics, official documentation, and validity checks in the DBMS Analyzer to construct a dependency graph.

\noindent \underline{\textbf{II. Data Generation Space.}} Towards true autonomy in data generation, we construct a structured generation space that enables the Data Agent to explore autonomously. Based on the dependency graph, we topologically order the nine core modules to progressively construct NL--SQL pairs. Then, a unified time-series meta-schema with queryable details is introduced to abstract heterogeneous schema models. To efficiently propagate dependency information across phases, we introduce lightweight Structured Phase Outputs (SPOs) and Flags as \textit{Dependency Signals}. A serialized \textit{Query Frame} is further used to maintain the intermediate query representation throughout generation. Finally, we define a \textit{Global Context} and combine it with the above structures to form the space state, whose transitions are driven by Agent interactions.

\noindent \underline{\textbf{III. Autonomous Agent Exploration.}} To enable autonomous fine-grained control over the generation process and exploration of the data space, we develop a Data Agent with orchestration, decision-making, and generation capabilities. At each exploration step, we first construct the generation state from the data space and the dependency information. Next, we design a Skill Repository comprising dozens of Skills that support diverse autonomous capabilities. A Semantic Decision Agent \(\mathcal{M}_D\) then selects a Skill \(\kappa\) and leverages it to determine an action \(a\). The action \(a\) drives exploration through semantic decisions and tool execution, while dynamically updating the set of eligible subsequent Skills. Based on the action, the Query Generation Agent \(\mathcal{M}_G\) is activated as needed to generate a \(\Delta Query\) for the corresponding query component under the current semantics. The Agent's operations constitute the Exploration Outcome, which is then used to update the space environment accordingly. The Data Agent continues exploration until all modules are completed, generating a batch of queries. In addition, we design a memory mechanism which is detailed later.

\noindent \underline{\textbf{IV. Online Adaptation.}} Online adaptation enables the Agent to continuously learn and improve, as illustrated by the red arrows in Figure~\ref{fig:framework}. To construct a high-quality benchmark, we guide the Agent from multiple perspectives, including semantic validity, coverage and diversity, and executability. Therefore, we propose three training-free online adaptation mechanisms. The \textit{Semantic-Aware Adaptation} uses rule-based and model-based validators to assess semantic consistency and guide Skill selection. The \textit{Distribution-Aware Adaptation} provides decision guidance through action priors derived from the distribution of validated data. The \textit{Execution-Aware Adaptation} leverages TSDB execution feedback to refine query generation. At step \(t\), the Data Agent leverages adaptation knowledge accumulated up to step \(t-1\), while the knowledge generated at step \(t\) is used to guide subsequent exploration.

\noindent \underline{\textbf{V. Temporal-Aware Hierarchical Memory.}} TSSQL data inherently exhibit high structural repetition but low instance-level overlap. 
To address this efficiency challenge, we extract temporal patterns from the raw state for efficient storage and retrieval. A hierarchical memory is then introduced, with two levels tailored to the decision process and fine-grained nested operator-chain generation, respectively. Before Agent activation, the system retrieves memory using a matching algorithm. A successful hit bypasses the computationally intensive Agent invocation and directly returns the result. Finally, we design an Adaptive Memory Management mechanism to dynamically update the memory. This memory substantially improves system efficiency while achieving a favorable trade-off between efficiency and generation performance.

\section{Dependency-Aware Generation Space}
\label{sec:4}

\subsection{Dependency Modeling}
Given an industrial TSDB schema $\mathcal{S}$, comprehensive TSSQL generation requires systematically exploring a vast query space spanning diverse temporal intents. Existing LLM-based approaches~\cite{TACO,OmniSQL,TS-NL2SQL} typically rely on seed examples, templates, or manually crafted instructions to guide generation. However, both paradigms remain anchored to human-specified starting points and rely on loosely constrained generation for expansion. Consequently, they lack an explicit data-centric modeling of the query space, limiting data coverage. Therefore, we systematically extract the syntactic and semantic constraints of TSSQL and construct a dependency directed acyclic graph (DAG) to explicitly model the implicit relationships among query components.

First, we model the TSSQL syntax space to enable valid generation and comprehensive exploration. An intuitive approach is to construct the syntax based on the Abstract Syntax Tree (AST). However, a complete AST is overly complex, introducing substantial overhead and limiting practical feasibility. Prior work~\cite{AST,HIER} explored SQL generation with lightweight customized ASTs, but these methods largely predate the LLM era, as tree-structured ASTs are poorly suited to serialized generation. To this end, we analyze the formal TSSQL grammar by traversing its production rules from the query-statement production and grouping related non-terminals according to the query semantics they govern. Productions describing the same query component are consolidated, while auxiliary lexical rules are omitted. This process yields nine core syntactic modules $V=\{\mu_{\mathrm{select}}, \mu_{\mathrm{from}}, ... , \mu_{\mathrm{result}}\}$, which capture the major structural dimensions of TSSQL queries. 

With these syntactic modules, we can comprehensively model all components of TSSQL. However, organizing them remains challenging, as syntactic order does not necessarily align with semantic reasoning. For example, although \textit{SELECT} appears first syntactically in a SQL statement, semantically its valid functions and target time series can only be determined after the query scope and grouping structure are established. We therefore first identify the semantic task of each syntactic module, including its query decision and corresponding choices. Specifically, for each module $\mu_i\in V$, we introduce a set of semantic decision variables $X_i=\{x_{i,1},x_{i,2},\ldots,x_{i,m_i}\},$ where each variable $x_{i,k}$ is associated with a candidate action space $\mathcal A_{i,k}$. We then define a dependency constraint $r(x_{i,k},x_{j,l})$ to exist whenever the resolved decision of $x_{i,k}$ in module $\mu_i$ affects the decision of $x_{j,l}$ in another module $\mu_j$. If $r(x_{i,k},x_{j,l})$ explicitly prohibits an action a of $x_{j,l}$, we formulate a binary validity function $\phi_r(x_{i,k},a)\in\{0,1\}$ and include $x_{i,k}$ in $\mathbf{x}_{j,l}^r$. The feasible action space of $x_{j,l}$ is then defined as:
\begin{equation}
\mathcal A_{j,l}^{\,r}(\mathbf{x}_{j,l}^r) = \left\{ a\in\mathcal A_{j,l} \mid \phi_r(\mathbf{x}_{j,l}^r,a)=1 \right\},
\end{equation}
where $\mathbf{x}_{j,l}^r$ denotes the preceding decision variables that restrict the action space of $x_{j,l}$, and $\phi_r=1$ indicates that the action is valid. Based on these dependency constraints, we define a dependency from module $\mu_i$ to $\mu_j$ whenever $r(x_{i,k},x_{j,l})$ exists, denoted by:
\begin{equation}
E_{ij} = \left\{ r \,\middle|\, \exists\, x_{i,k}\in \mathbf{x}_{j,l}^{r}, \; x_{j,l}\in X_j \right\}.
\end{equation}
Accordingly, we construct a module-level dependency DAG $\mathcal{G}=(V,E)$ by treating the syntactic modules in $V$ as vertices and adding a directed edge $(\mu_i,\mu_j)$ whenever $E_{ij}\neq\emptyset$.

The constraints $r$ are derived from three complementary sources. We first analyze the temporal semantics of TSSQL to identify ordering and compatibility constraints among temporal decisions. We then extract explicit cross-module restrictions from the official documentation, and further inspect the semantic validity checks implemented in the DBMS Analyzer to recover constraints enforced by the DBMS. The extracted rules are normalized into the predicate form defined above and collected as $R_c = R_T \cup R_D \cup R_A$. We further validate and consolidate the candidate constraints $R_c$, yielding the final set $R$ used to construct the module-level dependencies. The resulting dependency graph is shown in Figure~\ref{fig:framework}.


\subsection{Generation Space Construction}
Based on the dependency graph, we further model the generation process as a data space, providing the Data Agent with a perceivable and explorable environment. Specifically, given a TSDB schema $\mathcal{S}$, we construct a structured generation space, denoted by $\Omega(\mathcal S) = \left( \mathcal P,\, \widehat{\mathcal S},\, \Psi,\, \mathcal F,\, \Theta,\, \mathcal Q,\, \delta \right)$. Next, we describe each component in detail.

\textit{Generation Phases $\mathcal P$:} Given the dependency DAG $\mathcal G=(V,E)$, we topologically order the syntactic modules as $\pi=\operatorname{TopoSort}(\mathcal G).$ With initialization and finalization phases, we define the space backbone as $\mathcal P=(P_0,P_{\pi_1},P_{\pi_2},\ldots,P_{\pi_9},P_f)$, where each $P_{\pi_i}$ corresponds to the syntactic module $\mu_{\pi_i}$. Each phase $P_i$ contains its decision variables $X_i$ and action spaces $\{\mathcal A_{i,k}\}_{k=1}^{m_i}$ for agent interaction.

\textit{Unified TS-Schema $\widehat{\mathcal S}$:} To accommodate large-scale and heterogeneous TSDB schemas, we normalize the raw schema into a unified representation $\widehat{\mathcal S}=(\mathcal S^{M},\mathcal S^{D})$, where $\mathcal S^{M}$ provides a concise schema-level overview, while $\mathcal S^{D}$ maintains fine-grained schema details that can be efficiently retrieved on demand.

\textit{Structured Phase Output (SPO) $\Psi$:} SPOs provide lightweight propagation of cross-phase dependencies. For an edge $(\mu_i,\mu_j)\in E$, we extract the associated dependencies $r(x_{i,k},x_{j,l})$. If the decision result of $x_{i,k}$ semantically affects $x_{j,l}$, we record this result in the SPO $\Psi$. For each phase $P_j$, its input SPO is aggregated from the predecessor phases: $\Psi_j^{\mathrm{in}} = \bigoplus_{P_i\in \operatorname{Pred}(P_j)} \Pi_{ij}(\Psi_i),$ $\Pi_{ij}$ projects the dependency-relevant information from $\Psi_i$ for $P_j$, and $\oplus$ denotes aggregation over multiple predecessors.

\textit{Dependency Flags $\mathcal F$:} Flags encode explicit cross-phase restrictions. Specifically, if $r(x_{i,k},x_{j,l})$ induces a validity function $\phi_r(x_{i,k},a)$, the corresponding restriction is recorded in $\mathcal F$.

\textit{Global Context $\Theta$:} $\Theta$ maintains global context and shared system states, such as temporal information and runtime settings.

\textit{Query Frame $\mathcal Q$:} The Query Frame maintains a serialized intermediate representation of the partially constructed query, denoted as $Q=\operatorname{Serialize}\!\left(\{X_i\}_{i=1}^9,\mathcal O_T\right),$ which records the generated results of semantic decision variables $X_i$ together with the intermediate expressions constructed from the operator space $\mathcal O_T$.

\textit{Transition $\delta$:} At step $t$, given a resolved action $a_t\in\mathcal A_{i,k}$ for decision variable $x_{i,k}$, we define the transition as $\Omega_{t+1}=\delta(\Omega_t,a_t),$ where the relevant structures in $\Omega_t$ are updated accordingly.

\section{Data Agent}
\label{sec:5}

\subsection{Skill-Orchestrated Agent Architecture}
The generation space defines the boundary and structure of data construction, while raising a further challenge: how to effectively explore this space? Existing approaches~\cite{OmniSQL,SQL-Factory,TACO} typically use predefined workflows that assign LLMs or agents fixed roles, such as generation, augmentation, or verification. The generation trajectory is determined by the framework, limiting autonomous decision-making and exploration. To address this challenge and endow the Agent with data autonomy beyond monolithic prompting, we abstract dozens of fine-grained capabilities into Skills organized in phase-specific repositories. Specifically, we define two primary types of Skills: decision Skills and generation Skills. A decision Skill is represented as $\kappa^{D} = \left\langle x_{i,k},\mathcal A_{i,k},\rho_D\right\rangle,$ where $x_{i,k}$ is the target semantic decision variable, $\mathcal A_{i,k}$ is action space, and $\rho_D$ denotes the Skill execution policy. Given the current generation state $\Omega$, the decision Skill $\kappa^{D}$ produces an action $a=\rho_D\!\left(\Omega,x_{i,k},\mathcal A_{i,k}^{r}\right), a\in\mathcal A_{i,k}^{r}.$ Based on a resolved action $a$, the generation Skill is denoted as $\kappa^{G}=\langle a,\rho_G\rangle$, where $\rho_G$ produces an incremental intermediate representation $\Delta Q=\rho_G(\Omega,a)$. These Skills provide complementary capabilities, such as planning, reasoning, and instantiation, to support autonomous generation. Finally, each phase $P_i$ is associated with a dedicated Skill repository $\mathcal K_i= \left( \kappa_1^{D},\ldots,\kappa_m^{D}, \kappa_1^{G},\ldots,\kappa_n^{G} \right),$ containing Skills specialized for that phase.

Next, we decouple data exploration into two specialized agents: the \textit{Semantic Decision Agent} \(\mathcal M_D\) and the \textit{SQL Generation Agent} \(\mathcal M_G\). The former autonomously navigates the data generation space through semantic reasoning, while the latter incrementally constructs queries based on these decisions. Specifically, $\mathcal M_D$ makes semantic decisions such as selecting temporal patterns and grouping paradigms, guided by semantic context and decision Skill specifications. In turn, $\mathcal M_G$ constructs and updates the intermediate query representation by incrementally instantiating concrete query elements according to generation Skill specifications. The collaboration between the two agents prevents syntactic generation from directly determining high-level query semantics and enables semantic decisions to be explicitly maintained and constrained.



\begin{algorithm}
    \small
    \SetKwInOut{Input}{Input}
    \SetKwInOut{Output}{Output}

    \Input{
    state $\Omega'_t$; raw time series $TS$;
    Agents $\mathcal M_D$, $\mathcal M_G$;
    Skills: \{SeriesExpr Extension $\kappa_1^D$;
    AggExpr Extension $\kappa_2^D$;
    AGG Selection $\kappa_3^D$;
    TSGF Selection $\kappa_4^D$;
    Operator Selection $\kappa_5^D$;
    Expr Instantiation $\kappa_1^G$\}
    }

    \Output{SeriesExpr set $\mathcal C_S$; AggExpr set $\mathcal C_A$.}

    \caption{Compositional Operator-Chain Construction}
    \label{alg:operator_chain}

    $\mathcal C_S \gets TS,\quad
     \mathcal C_A \gets \varnothing$\;

    $r_t \gets
    \mathcal M_D(\Omega'_t,
    \{\kappa_1^D,\kappa_2^D,
    \kappa_1^D\!\rightarrow\!\kappa_2^D\})$\;
    
    \If{$r_t \in
    \{\kappa_1^D,\kappa_1^D\!\rightarrow\!\kappa_2^D\}$}{
        \While{\textit{SeriesExpr Construction} is not stopped}{
            $a_t \gets
            \mathcal M_D(\Omega'_t,\kappa_1^D)$\;
    
            $\kappa^D \gets
            \kappa_4^D$ if TSGF, otherwise $\kappa_5^D$
            according to $a_t$\;
    
            $(\Omega'_t,\mathcal C_S)
            \gets
            Exploration(\Omega'_t,\mathcal M_D,\mathcal M_G,
            \kappa^D,\kappa_1^G,\mathcal C_S)$\;
        }
    }
    
    \If{$r_t \in
    \{\kappa_2^D,\kappa_1^D\!\rightarrow\!\kappa_2^D\}$}{
    
        $(\Omega'_t,\mathcal C_A)
        \gets
        Exploration(\Omega'_t,\mathcal M_D,\mathcal M_G,
        \kappa_3^D,\kappa_1^G,\mathcal C_A)$\;
    
        \While{\textit{AggExpr Construction} is not stopped}{
            $a_t \gets
            \mathcal M_D(\Omega'_t,\kappa_2^D)$\;
    
            $\kappa^D \gets
            \kappa_4^D$ if TSGF, otherwise $\kappa_5^D$
            according to $a_t$\;
    
            $(\Omega'_t,\mathcal C_A)
            \gets
            Exploration(\Omega'_t,\mathcal M_D,\mathcal M_G,
            \kappa^D,\kappa_1^G,\mathcal C_A)$\;
        }
    }
    \Return $\mathcal C_S, \mathcal C_A$\;

\end{algorithm}

\begin{algorithm}
    \small

    \SetAlgorithmName{Function}{function}{List of Functions}
    \renewcommand{\thealgocf}{}

    \SetKwInOut{Input}{Input}

    \Input{
    $\Omega'_t,\mathcal M_D,\mathcal M_G,
    \kappa^D,\kappa^G,\mathcal C$
    }

    \caption{Exploration}

    $f_t \gets
    \mathcal M_D(\Omega'_t,\kappa^D), \Omega'_t \gets \operatorname{Update}(\Omega'_t, f_t)$\;

    $\Delta Q_t \gets
    \mathcal M_G(\Omega'_t,\kappa^G)$\;

    $\mathcal C \gets
    \mathcal C \cup \{\Delta Q_t\}$\;

    $\Omega'_t \gets
    \operatorname{WriteBack}(\Omega'_t,\Delta Q_t)$\;

    \Return $(\Omega'_t,\mathcal C)$\;

\end{algorithm}


\subsection{Autonomous Agent Exploration}

In this section, we detail how the Data Agent autonomously analyzes and explores the data space. Starting from the initialization phase \(P_0\), the Agent iteratively explores each phase until all phases are completed. We next describe the exploration process.

\textit{State Preparation: }Given the current space state \(\Omega_t\), we construct the input state \(\Omega'_t\) by extracting the phase-relevant \(\Theta\), \(\mathcal Q\), and \(\widehat{\mathcal S}\). Next, according to the dependency DAG \(\mathcal{G}=(V,E)\), we retrieve the SPOs $\Psi$ and Flags $\mathcal F$. For example, phase \(P_4\) obtains dependency from its related phases \(P_1\), \(P_2\), and \(P_3\), as illustrated in Figure~\ref{fig:framework}. The relevant SPOs \(\Psi\) are incorporated into \(\Omega'_t\) to provide the semantic dependency context, while the Flags \(\mathcal F\) impose action masks on the action space of the available Skills.

\textit{Dynamic Skill Routing: } After preparing the input state, we derive the set of currently eligible candidate Skills \(\mathcal{K}'_{i,t}\) from the phase-specific Skill repository. Rather than following a predetermined Skill sequence, the \textit{Semantic Decision Agent} \(\mathcal{M}_D\) dynamically selects a Skill from the candidate set based on the current semantics and the input state \(\Omega'_t\): $\kappa_t^D = \mathcal M_D \left( \Omega'_t, \mathcal K'_{i,t} \right)$. The Skill execution result further updates the set of eligible subsequent Skills. 

\textit{Agent Exploration: }Our agent progressively explores the data space using Skills. Specifically, each decision Skill defines the currently admissible exploration actions \(\mathcal A_{i,k}^{r}\). The \textit{Semantic Decision Agent} \(\mathcal M_D\) then selects an action $a_t=\mathcal M_D(\Omega'_t,\kappa_t^{D})$ according to the current input state and the Skill policy $\rho_D$. Each action advances the exploration and refines the query semantics. Execution actions invoke the corresponding tools, while generation actions activate the \textit{SQL Generation Agent} \(\mathcal M_G\) to instantiate the current semantics. When invoked, the action \(a_t\) provides \(\mathcal M_G\) with the generation Skill \(\kappa_t^{G}\) required for the current step. The agent then produces the corresponding incremental query: $\Delta Q_t=\mathcal M_G(\Omega'_t,\kappa_t^{G}).$ Additionally, before either agent is activated, a memory mechanism is accessed to directly obtain the required result when available. Further details are provided in Sec.~\ref{sec:6}.

\textit{Outcome Update: }The decision \(a_t\), its execution result, and the generated \(\Delta Q_t\) are recorded in the \textit{Exploration Outcome} and subsequently used to update the corresponding global context, query frame, and dependency-related variables in the data space.

\begin{figure}
  \centering  
  \includegraphics[width=0.9\linewidth]{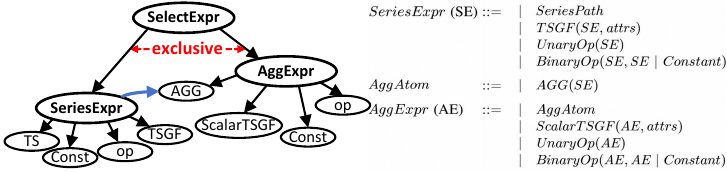}
  \caption{Recursive Operator Expression Structure}
  \label{fig:op}
\end{figure}

\noindent \textbf{Operator-Chain Exploration Algorithm: }TSDBs support operator chains over hundreds of temporal functions. To explore their semantics and combinations, we develop an iterative chain-construction algorithm based on the general exploration procedure.

Expressions in TSDBs include time-series generating functions ($TSGFs$) and aggregate functions ($AGGs$). The former produce timestamp-indexed data points, while the latter collapse points within a temporal segment to a single value. Due to their distinct computation granularity and cardinality, $SelectExpr$ has two mutually exclusive forms: $SeriesExpr$ and $AggExpr$. Additionally, the two expression types can be nested when the semantic and syntactic constraints are satisfied. Therefore, to automatically construct valid operator chains, we formulate a recursive expression grammar, as illustrated in Figure~\ref{fig:op}. A $SeriesExpr$ is initialized from raw time series \(TS\) and recursively extended with $TSGFs$, constants ($Const$), and operators ($op$). We then define $AggAtom$, which applies an \(AGG\) to an arbitrary $SeriesExpr$. An $AggExpr$ is built from $AggAtom$ and can be recursively extended with $Const$, $op$, and scalar-compatible $TSGFs$, excluding those that accept only sequence inputs.

As shown in Algorithm~\ref{alg:operator_chain}, we use operator-chain construction as an example to illustrate the exploration algorithm. We first initialize \(\mathcal C_S\) with raw time series and \(\mathcal C_A\) as an empty set. Next, $\mathcal M_D$ performs Skill routing to determine the generation trajectory. If $\mathcal M_D$ selects \(\kappa_1^D\), we iteratively construct the \textit{SeriesExpr} (lines 3-7 of Algorithm~\ref{alg:operator_chain}). Specifically, $\mathcal M_D$ iteratively invokes $\kappa_1^D$ to select the next \textit{SeriesExpr} extension action according to the grammar, then invokes $\kappa_4^D$ or $\kappa_5^D$ based on the selected action. Following this, we execute the \textbf{Exploration} procedure, where \(\mathcal M_D\) and \(\mathcal M_G\) jointly perform one instantiation step for \(\mathcal C_S\). Then, if \(\kappa_2^D\) is selected, we first construct an \textit{AggAtom} using \(\kappa_3^D\) and subsequently extend the \textit{AggExpr} iteratively in a manner analogous to \textit{SeriesExpr} (lines 8-13 of Algorithm~\ref{alg:operator_chain}). 

\textbf{Exploration}: within each exploration step, \(\mathcal M_D\) first invokes the input decision Skill \(\kappa^D\) to select a function or operator, including decisions on the function source, functional family, and concrete function. Subsequently, after incorporating the decision result into \(\Omega'_t\), \(\mathcal M_G\) invokes \(\kappa^G\) to instantiate an expression as the incremental query \(\Delta Q_t\). The resulting expression is then added to the corresponding set \(\mathcal C\), and the relevant state in \(\Omega'_t\) is updated accordingly.

\subsection{Online Adaptation}

Although the Data Agent can autonomously explore the data space, it still lacks the capacity for autonomous learning. High-quality TSSQL generation requires broad coverage, diversity, validity, and semantic plausibility. To enable efficient self-improvement of the Data Agent across these dimensions, we develop a training-free online adaptation mechanism that leverages exploration feedback through three complementary strategies, as detailed below.

\noindent \textbf{Distribution-aware adaptation: }Coverage and diversity are key dimensions of data generation quality. Existing approaches~\cite{Text2sql-Flow, TACO, SQL-Factory} typically improve diversity through offline augmentation by transforming query skeletons, templates, or existing queries. In contrast, we introduce online distribution-aware adaptation guided by the evolving data, enabling the Agent to autonomously improve its decision-making and explore underrepresented regions. For each semantic decision Skill $\kappa$, we track $N_{\kappa,t}(a)$, the number of valid instances generated by each action $a\in\mathcal A_\kappa$. The empirical action distribution at step $t$ is: 
\begin{equation}
    p_{\kappa,t}(a)
    =
    \frac{N_{\kappa,t}(a)+\epsilon}
    {\sum_{a'\in\mathcal A_\kappa}
    \left(N_{\kappa,t}(a')+\epsilon\right)},
\end{equation}
where $\epsilon$ is a smoothing factor. Given a target distribution \(p'_\kappa(a)\), which may be derived from real-world scenarios or user preferences, we quantify the distribution deficit of each action as $D_{\kappa,t}(a) = \frac{\left[p'_\kappa(a)-p_{\kappa,t}(a)\right]_{+}}{p'_\kappa(a)},$
where $[x]_{+}=\max(x,0)$. Larger deficits indicate greater underrepresentation in the generated data. We derive an action prior over the current feasible action set $\mathcal A_{\kappa,t}^{r}$ from these deficits and provide it to \(\mathcal M_D\) as online distributional knowledge:
\begin{equation}
    \varrho_{\kappa,t}(a)
    =
    \frac{
        \exp(D_{\kappa,t}(a)/\tau)
    }{
        \sum_{a'\in\mathcal A_{\kappa,t}^{r}}
        \exp(D_{\kappa,t}(a')/\tau)
    },
\end{equation}
where $\tau$ controls the strength of distribution-aware guidance. The action prior guides the Agent toward underexplored decisions while preserving semantic validity. 

\noindent \textbf{Execution-aware adaptation: }Despite our fine-grained modeling of TSSQL syntax and semantics, the inherent uncertainty of LLMs may still cause generated queries to violate DBMS-specific constraints. Therefore, we design an execution-aware adaptation mechanism that continuously distills DBMS execution feedback into reusable generation knowledge, enabling $\mathcal M_G$ to progressively improve generation quality. Given the complete query \(q_t\) generated by the Agent, we execute it on the database $\operatorname{Exec}(q_t,\mathcal S)\rightarrow (y_t,e_t),$
where \(y_t\) denotes the execution result and \(e_t\) captures the DBMS feedback, such as the status code and error message. $\mathcal M_G$ then repairs the query using execution feedback and historical experience, $q'_t=\mathcal M_G(q_t,e_t,\mathcal H_t)$, where $\mathcal H_t$ denotes experience in the execution knowledge base. After repair, we distill the repair experience into a reusable rule $h_t=\operatorname{Distill}(e_t,q_t,q't)$ and add it to the knowledge base: $\mathcal H{t+1}=\mathcal H_t\cup{h_t}$. For subsequent generation, we incorporate relevant knowledge retrieved from $\mathcal H_t$ based on the current query structure and operator signatures into the generation context. If execution fails, we retrieve additional experience based on the error category to guide repair.

\noindent \textbf{Semantic-aware adaptation: }Although the dependency mechanism improves decision-making, it offers limited guidance for Skill routing. We therefore introduce semantic-aware adaptation to learn the state-dependent Skill suitability from online semantic feedback. A rule-based validator checks local semantic consistency at each routing step, while an LLM-based validator assesses overall Skill route consistency after generation. Let \(v_t^{\mathrm{r}}, v_t^{\mathrm{LLM}} \in \{0,1\}\) denote the two validation results, we define the semantic reward as \(r_t = v_t^{\mathrm{r}} \cdot v_t^{\mathrm{LLM}}\). We then compress $\Omega'_t$ into a lightweight semantic context $c_t$ and track the selection count $N_t(c_t,\kappa)$ and cumulative semantic reward $R_t(c_t,\kappa)=\sum r_t$. A routing score is computed as:
\begin{equation}
    S_t(\kappa)
    =
    \frac{R_t(c_t,\kappa)+1}
         {N_t(c_t,\kappa)+2}
    +
    \lambda
    \sqrt{
        \frac{\log\!\left(N_t(c_t)+2\right)}
             {N_t(c_t,\kappa)+1}
    },
\end{equation}
where $N_t(c_t)=\sum_{\kappa}N_t(c_t,\kappa)$ and $\lambda$ controls the
exploration strength. The resulting scores are incorporated into the Agent context as online routing knowledge to guide subsequent Skill selection.

\section{Temporal-Aware Memory}
\label{sec:6}

\begin{figure*}
  \centering  
  \includegraphics[width=0.79\linewidth]{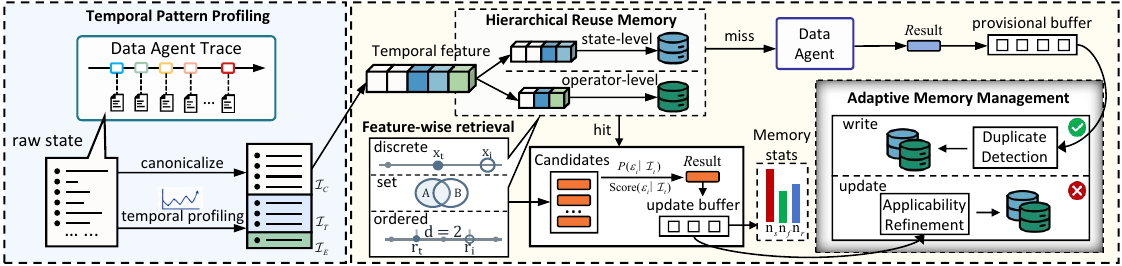}
  \caption{Overview of the Temporal-Aware Hierarchical Memory Mechanism}
  \label{fig:memory}
\end{figure*}


Temporal queries essentially instantiate recurring temporal semantics over heterogeneous schema entities. The same local temporal semantic structure may appear across industrial scenarios, while their schemas and temporal characteristics, such as sampling granularity, can differ substantially. This gives rise to a structure-instance asymmetry between reusable temporal structures and heterogeneous data instances. However, the reuse of these structures depends not only on schema compatibility but also on the temporal characteristics and semantics of the underlying data. To address this challenge, we propose a temporal-aware hierarchical memory that efficiently models temporal characteristics and enables the reuse of generation experience across instances. As illustrated in Figure~\ref{fig:memory}, we next detail the memory mechanism.

\noindent \textbf{Temporal Pattern Profiling: }To identify reusable experience, we need to extract temporal characteristics from the data space. The raw state $\Omega_t$ is unsuitable for direct matching due to substantial redundancy. We therefore extract a temporal profile to capture the current data characteristics and temporal semantics of the data space. Specifically, we first canonicalize \(\Omega_t\) as $\mathcal{I}_C=\operatorname{Canon}(\Omega_t),$ where \(\operatorname{Canon}\) prunes the dependency information \(\Psi\), \(\mathcal F\), and query representation \(\mathcal Q\), and maps their state values into matchable ordered sequences, sets, and atomic values. Next, we design temporal profiling to extract a low-dimensional, instance-agnostic representation of the temporal context. For the \(m\) time series across \(n\) devices in the current space state, let \(I_i=[l_i,r_i]\) denote the valid interval of the \(i\)-th series, \(D=[L,R]\) the overall data interval with \(L=\min_i l_i\) and \(R=\max_i r_i\), \(Q=[a,b]\) the query range, and \((\alpha,\beta)\) the window parameters. We compute:
\begin{equation}
    \mathcal{I}_T= \left\langle \sigma_1(n,m,\{I_i\}), \sigma_2(R-L), \sigma_3\!\left(\frac{|Q|}{|D|}\right), \ldots, \sigma_k(\alpha,\beta) \right\rangle,
\end{equation}
which characterize temporal patterns, data span, query coverage, window configuration, and other temporal properties. Each $\sigma_i(\cdot)$ discretizes a temporal characteristic into a compact representation, abstracting instance-specific values while retaining temporal information relevant to experience reuse. For expression-related states, we additionally extract an expression profile \(\mathcal{I}_E\), which captures structural properties of the current expression, such as its operator composition and input/output characteristics. Finally, we combine the above features into a unified temporal profile $\mathcal{I}=\left\langle \mathcal{I}_C,\mathcal{I}_T,\mathcal{I}_E \right\rangle$, which serves as the key for subsequent memory retrieval.

\noindent \textbf{Hierarchical Reuse Memory: }Reusable decision and generation experiences are associated with distinct temporal characteristics. We therefore design a hierarchical memory architecture consisting of \textit{state-level memory} and \textit{operator-level memory}, which store semantic decision results for $\mathcal{M}_D$ and operator expressions for $\mathcal{M}_G$, respectively. The current temporal profile $\mathcal I_t$ is used to retrieve the memory. A valid match enables direct result reuse without Agent invocation; otherwise, the Agent is invoked, and its experience may be stored after validation. Memory matching is based on a similarity function \(\operatorname{Sim}(\mathcal I_t,\mathcal I_i)\) that compares the current temporal profile with each stored memory entry. Categorical features use exact or neighboring-category matching, set-valued features use set similarity, and ordered features use distance-based soft matching. Entries above the similarity threshold form the retrieved candidate set $\mathcal{H}_t$. For each memory entry $\varepsilon_i$, we maintain its successful reuse count $N_i^{s}$, failed reuse count $N_i^{f}$, and total reuse count $N_i^{r}$. Then, its confidence is estimated as $\operatorname{Conf}(\varepsilon_i) = \frac{N_i^{s}+1}{N_i^{s}+N_i^{f}+2}.$
When multiple entries are matched, we score them by effectiveness and frequency:
\begin{equation}
    \operatorname{Score}(\varepsilon_i\mid\mathcal I_t)
    =
    \frac{
        \operatorname{Sim}(\mathcal I_t,\mathcal I_i)
        \cdot
        \operatorname{Conf}(\varepsilon_i)
    }{
        1+\lambda N_i^{r}
    },
\end{equation}
where $\lambda$ controls the penalty on repeatedly reused entries.
Based on the retrieval scores, the system retains the top-$K$ entries from the retrieved candidate set $\mathcal H_t$ and constructs a normalized selection distribution:
\begin{equation}
    P(\varepsilon_i\mid\mathcal I_t)
    =
    \frac{
        \operatorname{Score}(\varepsilon_i\mid\mathcal I_t)
    }{
        \sum_{\varepsilon_j\in\operatorname{TopK}(\mathcal H_t)}
        \operatorname{Score}(\varepsilon_j\mid\mathcal I_t)
    },
\end{equation}
The system then samples one memory entry from this distribution as the retrieval result.

\noindent \textbf{Adaptive Memory Management: } To maintain the reliability and applicability of the reuse memory, we dynamically manage memory entries through admission, update, refinement, and eviction. New experiences produced by the Data Agent are first stored in a provisional buffer and admitted into memory only after validation and redundancy checking. When a memory entry is retrieved, its reuse outcome is temporarily recorded in an update buffer; after validation, the corresponding success, failure, and reuse statistics are updated. If reuse fails due to temporal incompatibility, we refine the entry's applicability as $\mathcal I_i \leftarrow \operatorname(\mathcal I_i,\mathcal I_f)$, where $\mathcal I_f$ denotes the failed reuse profile. Finally, an entry $\varepsilon_i$ is removed when its failure ratio exceeds a threshold $\frac{N_i^{f}}{N_i^{s}+N_i^{f}}> \gamma .$
This process allows the memory to continuously absorb validated experience, refine reuse conditions, and discard unreliable entries during generation.


\section{Experiments}
\label{sec:7}

\subsection{Experimental Setup}

\begin{table*}[t]
\centering
\caption{Statistics of different NL2SQL datasets.}
\label{tab:dataset_statistics}
\scalebox{0.8}{
\begin{tabular}{l|ccccccccc}
\toprule
\textbf{Method}
& \shortstack{\textbf{\#}\\\textbf{Example}}
& \shortstack{\textbf{\#}\\\textbf{DB}}
& \shortstack{\textbf{\# Table}\\\textbf{/ DB}}
& \shortstack{\textbf{\# Col}\\\textbf{/ DB}}
& \shortstack{\textbf{\# Tok.}\\\textbf{/ SQL}}
& \shortstack{\textbf{\# Unique}\\\textbf{Func.}}
& \shortstack{\textbf{Build}\\\textbf{Mode}}
& \shortstack{\textbf{TSDB}\\\textbf{-native}}
& \shortstack{\textbf{Time-series}\\\textbf{Analytics}} \\
\midrule

WikiSQL~\cite{WikiSQL}                   & 80,654 & 26,521 & 1.00   & 6.34    & 12.2  & 5   & Hybrid    & \xmark & \xmark \\
Spider 1.0~\cite{Spider}                 & 10,181 & 200    & 5.11   & 26.82   & 18.5  & 5   & Manual    & \xmark & \xmark \\
ScienceBenchmark~\cite{ScienceBenchmark} & 5,031  & 3      & 16.67  & 86.67   & 15.6  & 5   & Hybrid    & \xmark & \xmark \\
BIRD~\cite{BIRD}                         & 12,751 & 95     & 7.64   & 54.56   & 30.9  & 24  & Manual    & \xmark & \xmark \\
Spider 2.0-Snow~\cite{Spider2}           & 547    & 152    & 52.63  & 812.1   & 161.8 & 106 & Manual    & \xmark & \xmark \\
BIRD-History~\cite{BIRD-History}         & 1,393  & 11     & 6.8    & 72.5    & 80.9  & 22  & Automatic & \xmark & \xmark \\
TACO-Beijing~\cite{TACO}                 & 6,705  & 24     & 92.1   & 1458    & 14.6  & 9   & Automatic & \xmark & \xmark \\
NL2InfluxDB~\cite{TS-NL2SQL}             & 1,106  & 1      & -      & -       & 18    & -   & Automatic & \cmark & \cmark \\

\midrule

\textbf{TERN}                            & 13,182 & 32     & 92.19  & 2313.34 & 38.08  & 127 & Automatic & \cmark & \cmark \\

\bottomrule
\end{tabular}
}
\end{table*}

\noindent \textbf{Datasets.} For \textit{TSDB} data, we collect 32 large-scale time-series datasets from real-world industrial scenarios and construct 32 \textit{TSDB} instances. The databases average 92.19 devices and 2,313.34 measurements, with respective maxima of 963 and 16,920. On average, each database contains 4.5 million data points spanning 797.73 days.

\noindent \textbf{Baseline methods.} We evaluate the \BenchName{} benchmark and our \modelName{} method with two categories of baselines.

\noindent (1) \textbf{NL2SQL Baselines.} We extensively evaluate \BenchName{} with fine-tuned LLMs, frontier LLMs, and mainstream NL2SQL methods: 

\begin{itemize}[leftmargin=10pt]
    \item \textbf{Fine-tuned LLMs}: To assess transferability to TSSQL and ensure comparability, we perform supervised fine-tuning (SFT) on representative open-source LLMs, including Meta-Llama-3.1-8B~\cite{Llama}, GLM-4-9B~\cite{glm}, the Qwen3 family~\cite{qwen} (Qwen3-8B, Qwen3-30B-A3B, and Qwen3.8-27B), and code model Qwen2.5-Coder-7B~\cite{qwen_coder2.5}.
    \item \textbf{Frontier LLMs}: We evaluate representative proprietary LLM families on Text-to-TSSQL, including GPT (GPT-5.6-sol~\cite{gpt56} and GPT-6.1-sol~\cite{gpt61}), Qwen~\cite{qwen} (Qwen3.7-Max, Qwen3.8-Max, and Qwen3-Coder-480B), and DeepSeek-V4.1-Flash~\cite{deepseek}.
    \item \textbf{NL2SQL Methods}: We evaluate mainstream NL2SQL methods, including fine-tuned approaches such as CodeS~\cite{CodeS}, OmniSQL~\cite{OmniSQL}, and XiYan-SQL~\cite{XiYan-SQL}, as well as proprietary LLM-based methods such as DAIL-SQL~\cite{DAIL-SQL}, OpenSearch-SQL~\cite{OpenSearch-SQL}, DeepEye-SQL~\cite{DeepEye-SQL}, and NL2InfluxDB~\cite{TS-NL2SQL}.
\end{itemize}

\noindent (2) \textbf{NL-SQL Generation Baselines.} We compare \modelName{} with:

\begin{itemize}[leftmargin=10pt]
    \item \textbf{DeepSeek}~\cite{deepseek}: We use DeepSeek-V4.1-Flash as a reference baseline, augmented with RAG~\cite{RAG} over a TSSQL knowledge base.
    \item \textbf{OmniSQL}~\cite{OmniSQL}: This method proposes a four-stage LLM-driven data synthesis pipeline and employs semantic consistency selection and majority voting to select high-quality data.
    \item \textbf{SQL-Factory}~\cite{SQL-Factory}: This method proposes a multi-agent SQL generation framework with three collaborative teams for generating initial queries, expanding seed queries, and validation.
    \item \textbf{SemanticAgent}~\cite{SemanticAgent}: This method constructs a structured semantic knowledge base to guide data synthesis and employs knowledge-grounded diagnosis and iterative refinement.
    \item \textbf{TACO}~\cite{TACO}: This work proposes a three-stage data synthesis pipeline that employs grammar-based skeleton generation, schema-guided query instantiation, and CoT-based NL generation.
    \item \textbf{NL2InfluxDB}~\cite{TS-NL2SQL}: This work proposes a documentation-driven data synthesis pipeline for InfluxDB that integrates retrieval-augmented generation and knowledge-graph-based verification.
\end{itemize}

\noindent \textbf{Benchmark Configuration.} To enable more fine-grained evaluation on \BenchName{}, we further organize the benchmark into three versions. \BenchName{}-Foundation covers basic TSSQL knowledge with relatively short, simple queries. \BenchName{}-Standard, the largest subset, targets moderately difficult NL2TSSQL tasks. \BenchName{}-Analytical focuses on time-series analytical queries with more complex temporal operator chains. Each version is split into training, validation, and test sets at an 8:1:1 ratio. All collected benchmark instances are validated through LLM-based and manual verification.

\noindent \textbf{Evaluation Metrics.} For \BenchName{}, we adopt the widely used~\cite{BIRD, TACO, SemanticAgent} \textit{Execution Accuracy} (EX) and \textit{Successful Execution Rate} (SER). EX measures the proportion of predictions matching ground-truth execution results, while SER measures the proportion that execute successfully. For \modelName{}, we evaluate the coverage, diversity, and complexity. Following SQL-Factory~\cite{SQL-Factory}, we compute \textit{Hybrid Similarity} and the \textit{Vendi Scores} to evaluate sample similarity. The \textit{TSSQL Coverage} metric builds on SQL coverage criteria from prior work~\cite{coverage}, covering temporal, SQL-structural, and schema dimensions. The \textit{Hill diversity} metric further measures TSSQL diversity, defined as \(\mathrm{HD}=\exp\left(-\sum_{i=1}^{K}p_i\ln p_i\right)\), where \(p_i\) is the frequency of the \(i\)-th TSSQL feature category. Finally, following the SQL adaptation~\cite{Complexity} of Halstead measures~\cite{Halstead}, we define \textit{Halstead Complexity} as the geometric mean of the Halstead measure and expression depth. The Halstead measure is computed as \(\mathrm{H}(q)=\frac{n_1}{2}\cdot\frac{N_2}{n_2}\cdot\log_2(n_1+n_2)\), where \(n_1\), \(n_2\), and \(N_2\) denote the numbers of distinct operators, distinct operands, and total operand occurrences.

\noindent \textbf{Experimental Settings.} All experiments are implemented with PyTorch 2.6 and Python 3.11 on Ubuntu 26.04. Methods requiring training are trained using 4 NVIDIA A100 GPUs.

\subsection{Benchmark Statistics}

Table~\ref{tab:dataset_statistics} compares \BenchName{} with representative NL2SQL benchmarks, including established benchmarks such as WikiSQL~\cite{WikiSQL}, Spider~\cite{Spider}, ScienceBenchmark~\cite{ScienceBenchmark}, and BIRD~\cite{BIRD}, as well as recent challenging benchmarks, including Spider 2.0-Snow~\cite{Spider2}, BIRD-History~\cite{BIRD-History}, and TACO-Beijing~\cite{TACO}. We also include NL2InfluxDB~\cite{TS-NL2SQL}, a recent attempt at NL2SQL for time-series databases. However, its benchmark data are not publicly available, preventing complete statistics.

Our benchmark \BenchName{} contains 13,182 examples collected from 32 real-world industrial TSDBs. The schema scale is comparable to recent complex benchmarks such as Spider 2.0-Snow and TACO-Beijing. In particular, the industrial TSDBs contain numerous measurements, averaging 2,313.34 columns per database and up to 16,920. The SQL length is computed following Spider 2.0~\cite{Spider2}. Since \BenchName{} is not constructed to contain only highly complex queries, its SQL length is lower than that of benchmarks specifically targeting complex SQL, such as Spider 2.0-Snow and BIRD-History. Nevertheless, rich temporal syntax and complex expressions make its queries longer than those in other benchmarks. More importantly, \BenchName{} covers 127 unique time-series and aggregate functions, exceeding even Spider 2.0-Snow, which contains 106 unique functions.

In addition, \BenchName{} differs from existing benchmarks in its focus on TSDB-native queries and time-series analytics. Although NL2InfluxDB explores this direction with 1,106 automatically synthesized examples, it uses a single database with limited query and schema complexity and lacks publicly available data. To address this gap, we introduce \BenchName{}, providing large-scale coverage of TSDB-native analytical queries.

\begin{table*}[t]
\centering
\caption{Performance comparison on the TERN benchmark.}
\label{tab:main_results}
\scalebox{0.74}{
\begin{tabular}{l|CC|CC|CC|CC}
\toprule

\multirow{2}{*}{\textbf{Method}}
& \multicolumn{2}{c|}{\textbf{TERN-Foundation}}
& \multicolumn{2}{c|}{\textbf{TERN-Standard}}
& \multicolumn{2}{c|}{\textbf{TERN-Analytical}}
& \multicolumn{2}{c}{\textbf{Overall}} \\

\cline{2-3}\cline{4-5}\cline{6-7}\cline{8-9}

& \textbf{EX} & \textbf{SER}
& \textbf{EX} & \textbf{SER}
& \textbf{EX} & \textbf{SER}
& \textbf{EX} & \textbf{SER} \\

\midrule
\multicolumn{9}{c}{\textbf{Fine-tuned LLMs ($\leq$32B)}} \\
\midrule
Qwen3.8-27B(ICL)                & 9.17\% & 15.83\% & 4.34\% & 12.10\% & 2.45\% & 8.87\% & 5.33\% & 12.36\% \\
GLM-4-9B(SFT)                   & 58.61\% & 73.33\% & 29.22\% & 73.06\% & 21.71\% & 43.73\% & 36.44\% & 64.62\% \\
Meta-Llama-3.1-8B-Instruct(SFT) & 65.56\% & 75.83\% & 42.01\% & 74.89\% & 27.83\% & 51.07\% & 45.42\% & 68.27\% \\
Qwen2.5-Coder-7B-Instruct(SFT)  & 55.83\% & 75.55\% & 36.07\% & \textbf{83.79\%} & 27.52\% & 58.72\% & 39.91\% & 73.87\% \\
Qwen3-30B-A3B-Instruct(SFT)     & 69.17\% & 78.89\% & 40.64\% & 81.96\% & 29.05\% & 52.29\% & 46.40\% & 72.36\% \\
Qwen3.8-27B(SFT)                & \textbf{74.44\%} & 82.78\% & 45.89\% & 78.31\% & 29.97\% & 56.88\% & \textbf{50.40\%} & 73.51\% \\

\midrule
\multicolumn{9}{c}{\textbf{Frontier LLMs}} \\
\midrule

Qwen3.7-max                 & 16.94\% & 45.28\% & 7.08\% & 27.40\% & 11.31\% & 25.38\% & 11.46\% & 32.53\% \\
Qwen3.8-max                 & 18.06\% & 46.39\% & 6.39\% & 21.23\% & 8.87\% & 31.80\% & 10.84\% & 32.36\%  \\
Qwen3-Coder-480B            & 29.17\% & 55.00\% & 7.53\% & 29.45\% & 8.26\% & 27.52\% & 14.67\% & 37.07\% \\
DeepSeek-V4.1-Flash         & 54.72\% & 81.11\% & 31.05\% & 52.05\% & 22.94\% & 40.98\% & 36.27\% & 58.13\% \\
GPT-5.6-sol                 & 49.17\% & 71.94\% & 35.84\% & 63.01\% & 26.30\% & 39.45\% & 37.33\% & 59.02\% \\
GPT-6.1-sol                 & 68.61\% & \textbf{92.22\%} & \textbf{46.35\%} & 79.68\% & \textbf{30.89\%} & \textbf{59.63\%} & 48.98\% & \textbf{77.87\%} \\

\midrule
\multicolumn{9}{c}{\textbf{NL2SQL Methods}} \\
\midrule

CodeS-15B (SFT)             & 31.39\% & 51.94\% & 10.50\% & 28.31\% & 7.95\% & 18.04\% & 16.44\% & 32.89\% \\
OmniSQL-32B (SFT)           & 63.06\% & 76.67\% & 35.16\% & 78.31\% & 28.75\% & 56.27\% & 42.22\% & 71.38\% \\
XiYan-SQL-32B (SFT)         & 64.72\% & 78.61\% & 38.58\% & 80.82\% & 27.83\% & 53.52\% & 43.82\% & 72.18\% \\
DAIL-SQL                    & 11.39\% & 23.89\% & 6.62\% & 18.95\% & 5.81\% & 15.90\% & 7.91\% & 19.64\% \\
OpenSearch-SQL              & 30.83\% & 42.50\% & 8.22\% & 18.95\% & 3.06\% & 8.56\% & 13.96\% & 23.47\% \\
DeepEye-SQL                 & 42.50\% & 69.17\% & 11.64\% & 47.26\% & 8.87\% & 25.38\% & 20.71\% & 47.91\% \\

\bottomrule
\end{tabular}
}
\end{table*}

\subsection{Evaluation on Benchmark}

We comprehensively evaluate existing methods on \textbf{\BenchName{}} across three categories: Fine-tuned LLMs, Frontier LLMs, and NL2SQL Methods. As shown in Table~\ref{tab:main_results}, we draw the following observations:

\noindent (1) \textbf{Fine-tuned LLMs.} Without fine-tuning, these models perform poorly, with even Qwen3.8-27B achieving only 5.33\% overall EX. We therefore apply SFT, which substantially improves their EX and SER, supporting the effectiveness of our training corpus. Qwen3.8-27B achieves the highest EX across all three subsets, suggesting that larger models yield better performance. Qwen2.5-Coder-7B shows no clear advantage over similarly sized models, indicating limited benefits of code-oriented pretraining for TSSQL.

\noindent (2) \textbf{Frontier LLMs.} Frontier LLMs exhibit substantial performance variation. GPT-6.1 achieves the best EX and SER, outperforming GPT-5.6 by 11.65 pp in overall EX. In contrast, Qwen3.8 underperforms Qwen3.7, potentially due to excessive reasoning. These differences suggest varying degrees of transferability to TSSQL, making task-specific validation essential for model selection.

\noindent (3) \textbf{NL2SQL Methods.} SFT-based NL2SQL models show no clear advantage over general-purpose models, potentially due to the limited benefits of standard SQL pretraining for TSSQL. ICL-based methods perform relatively poorly, likely because their techniques designed for relational databases inadequately address TSSQL challenges. 

\noindent (4) \textbf{Cross-paradigm Comparison.} Comparing the strongest models from each category, Qwen3.8-27B achieves the highest EX on \BenchName{}-F, while GPT-6.1 performs better on \BenchName{}-S and \BenchName{}-A. This suggests that SFT is particularly effective for basic TSSQL queries, whereas GPT-6.1's stronger reasoning capabilities better support complex TSSQL queries and temporal analytics. 

\noindent (5) \textbf{Task-specific Performance Analysis.} Across all baselines, EX generally declines from \BenchName{}-F to \BenchName{}-S and \BenchName{}-A. Even the best model achieves less than one-third EX on \BenchName{}-A, highlighting the challenges of predicting analytical time-series queries.

\noindent (6) \textbf{Executability vs. Correctness.} Even the best methods achieve only around 75\% overall SER and 50\% EX, indicating that limited ability to generate executable TSSQL constrains EX accuracy.

\subsection{Error Analysis}

\begin{figure}
  \centering  
  \includegraphics[width=0.9\linewidth]{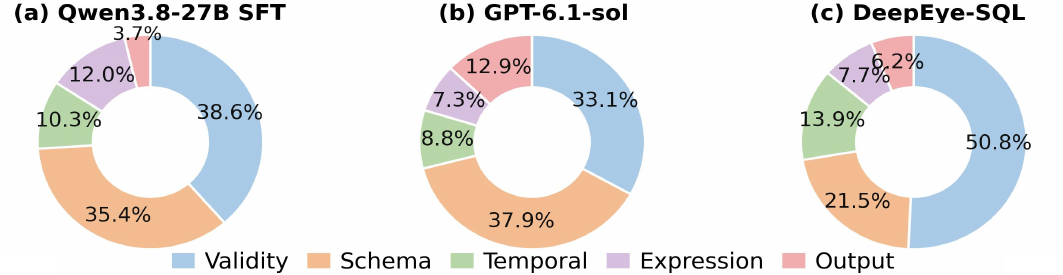}
  \caption{Statistics of Errors}
  \label{fig:error}
\end{figure}

To characterize failures, we evaluate the strongest baseline in each category on 300 randomly sampled examples from \BenchName{}. We analyze incorrect predictions and assign each a primary error category: \textit{Validity} (syntax or execution violations), \textit{Schema} (schema grounding or organization), \textit{Temporal} (temporal structure or operation), \textit{Expression} (function or computation), or \textit{Output} (output organization). Figure~\ref{fig:error} shows that Qwen3.8-27B has higher temporal and expression error shares than GPT-6.1-sol, suggesting weaker reasoning capabilities for complex temporal structures and expressions. GPT-6.1-sol has the lowest validity share but the highest schema and output shares, indicating potential benefits from dedicated schema linking and output organization mechanisms. DeepEye-SQL has the largest validity share, suggesting limited transferability to TSSQL.

\begin{figure}
  \centering  
  \includegraphics[width=0.87\linewidth]{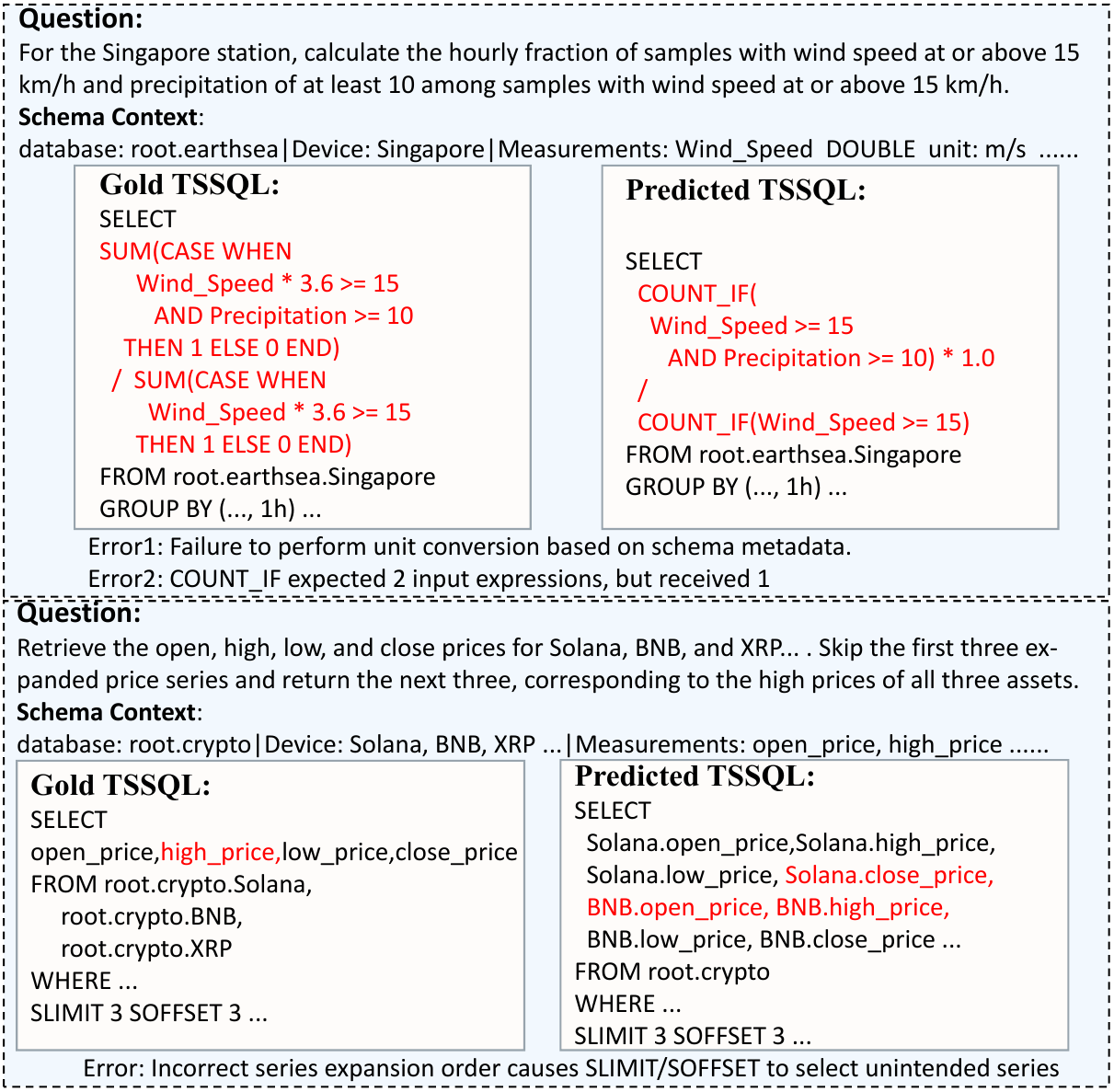}
  \caption{Examples of Prediction Errors}
  \label{fig:case_study}
\end{figure}

Figure~\ref{fig:case_study} presents concrete TSSQL examples and errors made by GPT-6.1-sol. In the first, the model omits the unit conversion implied by schema context and selects an inappropriate function. In addition, its \texttt{COUNT\_IF} calls also lack a required argument. In the second, it enumerates series by device rather than by measurement, changing their expansion order. Consequently, \texttt{SLIMIT 3 SOFFSET 3}, which skips the first three series and returns the next three, selects an incorrect subset (highlighted in red) instead of the three requested \texttt{high\_price} series, illustrating the challenges posed by specialized time-series schemas and output organization. 

\begin{figure*}
  \centering  
  \includegraphics[width=0.9\linewidth]{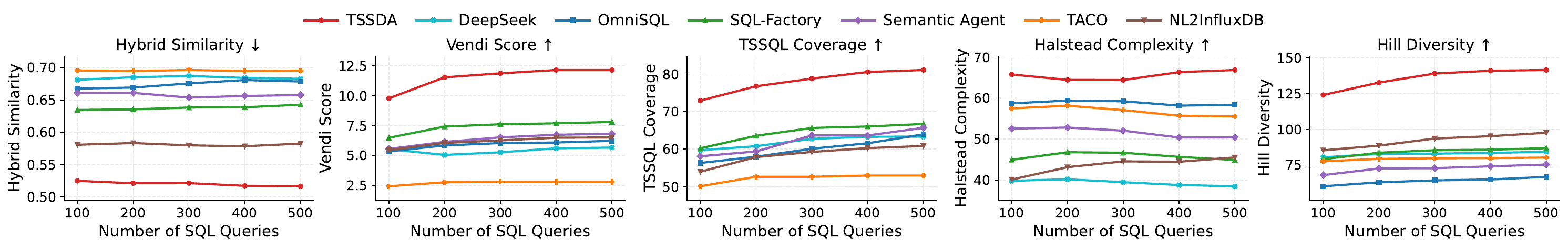}
  \caption{Scalability Comparison of Different Methods}
  \label{fig:sca}
\end{figure*}

\subsection{Effectiveness Comparison}

To evaluate generation effectiveness, we generate 500 executable samples using each baseline on the same TSDB. For a fair comparison, all baselines use DeepSeek-V4.1-Flash as the backbone model. In addition to the baselines, we evaluate three ablated variants of \modelName{}, removing semantic dependencies, online adaptation, and memory, respectively, to assess each component's contribution. Table~\ref{tab:effect} reports the results and we have the following observations:

\noindent (1) \textbf{Overall Effectiveness.} \modelName{} achieves the best performance across all metrics. It improves Vendi Score, TSSQL Coverage, and Hill Diversity by 55.80\%, 21.56\%, and 45.00\%, respectively, over their strongest baselines. These results demonstrate its effectiveness in generating diverse TSSQL queries with broad feature coverage.

\noindent (2) \textbf{Generation Diversity.} Hybrid Similarity and Vendi Score assess sample similarity, with \modelName{} achieving the best results on both. TACO performs worst, largely due to repeated instantiation of a limited skeleton set. \modelName{} also achieves the highest Hill Diversity, whereas OmniSQL records the lowest score, indicating concentrated feature reuse despite its complexity-oriented generation. Among the baselines, SQL-Factory achieves the highest Vendi Score, potentially benefiting from its diversity-oriented query rewriting. In contrast, NL2InfluxDB performs best on Hybrid Similarity and Hill Diversity, likely due to its balanced query-family sampling.

\noindent (3) \textbf{Query Complexity.} Among baselines, OmniSQL and TACO produce the most complex queries, consistent with their complexity-oriented designs, while DeepSeek produces the simplest. SQL-Factory and NL2InfluxDB also show relatively low complexity. This trend contrasts with the diversity results, potentially because complex queries share more common SQL structures. In contrast, \modelName{} achieves the best results in both complexity and diversity.

\noindent (4) \textbf{TSSQL Coverage.} \modelName{} achieves the highest coverage, benefiting from its broad generation space and highly autonomous Agent, which facilitate exploration of previously unexplored query features. In contrast, TACO performs worst, indicating that its limited skeleton set constrains generation beyond predefined patterns.


\noindent (5) \textbf{Ablation Analysis.} Removing adaptation causes the largest degradation on most metrics, highlighting its central role in effectiveness. Dependency removal also degrades performance, supporting the value of semantic constraints and dependency context. Removing memory yields only minor changes, including slight gains on some metrics, suggesting that memory, designed to improve efficiency through experience reuse, largely preserves effectiveness.



\begin{table}[t]
\centering
\caption{Effectiveness comparison of different methods.}
\label{tab:effect}
\scalebox{0.75}{
\renewcommand{\arraystretch}{1.0}
\begin{tabular}{l|YYYYY}
\toprule

\textbf{Method}
& \shortstack{\textbf{Hybrid}\\[-2pt]\textbf{Similarity}}
& \shortstack{\textbf{Vendi}\\[-2pt]\textbf{Score}}
& \shortstack{\textbf{TSSQL}\\[-2pt]\textbf{Coverage}}
& \shortstack{\textbf{Halstead}\\[-2pt]\textbf{Complexity}}
& \shortstack{\textbf{Hill}\\[-2pt]\textbf{Diversity}} \\

\midrule
DeepSeek-V4.1 & 0.6827 & 5.6563 & 63.3424 & 38.4834 & 84.2069 \\
OmniSQL & 0.6786 & 6.2234 & 63.9392 & 58.3691 & 66.7064 \\
SQL-Factory & 0.6426 & 7.8040 & 66.7029 & 44.9125 & 86.7652 \\
SemanticAgent & 0.6575 & 6.8108 & 65.6844 & 50.4266 & 75.3282 \\
TACO & 0.6953 & 2.7882 & 52.9698 & 55.5263 & 80.2348 \\
NL2InfluxDB & 0.5824 & 6.5023 & 60.8359 & 45.5195 & 97.5521 \\

\midrule
\textbf{TSSDA} & \textbf{0.5164} & 12.1590 & \textbf{81.0817} & 66.8761 & \textbf{141.4524} \\
w/o dependency & 0.5203 & 12.0060 & 79.6381 & 64.7948 & 136.7946 \\
w/o adaptation & 0.5580 & 10.1385 & 74.9015 & 65.5940 & 125.8635 \\
w/o memory & 0.5178 & \textbf{12.2656} & 80.9504 & \textbf{66.9497} & 141.2948 \\

\bottomrule
\end{tabular}
}
\end{table}

\subsection{Scalability Comparison}

To assess how generation effectiveness scales with dataset size, we increase the number of generated SQL queries from 100 to 500 in increments of 100 and track all five metrics for \modelName{} and the baselines. Observations derived from Figure~\ref{fig:sca} are as follows:

\noindent (1) As the dataset grows, \modelName{}'s Hybrid Similarity decreases overall, whereas most baselines remain stable or increase. Its Vendi Score, TSSQL Coverage, and Hill Diversity remain the highest and generally increase with dataset size, while most baselines show smaller gains. After an initial decline, \modelName{}'s complexity recovers and increases overall, while most baselines remain stable or decline. These trends suggest that \modelName{} maintains superior effectiveness as generation scales.

\noindent (2) From 100 to 500 queries, \modelName{} achieves the largest absolute gains in Vendi Score, TSSQL Coverage, and Hill Diversity, increasing by 2.37, 8.16, and 17.48, respectively. The corresponding largest baseline gains are 1.33 (SQL-Factory), 7.62 (OmniSQL), and 12.28 (NL2InfluxDB). These gains are achieved despite \modelName{} starting from substantially higher scores.

\noindent (3) Across all evaluated dataset sizes, \modelName{} consistently achieves the best results on all five metrics, demonstrating sustained effectiveness as generation scales.


\subsection{Efficiency Comparison}

\begin{figure}
  \centering  
  \includegraphics[width=0.98\linewidth]{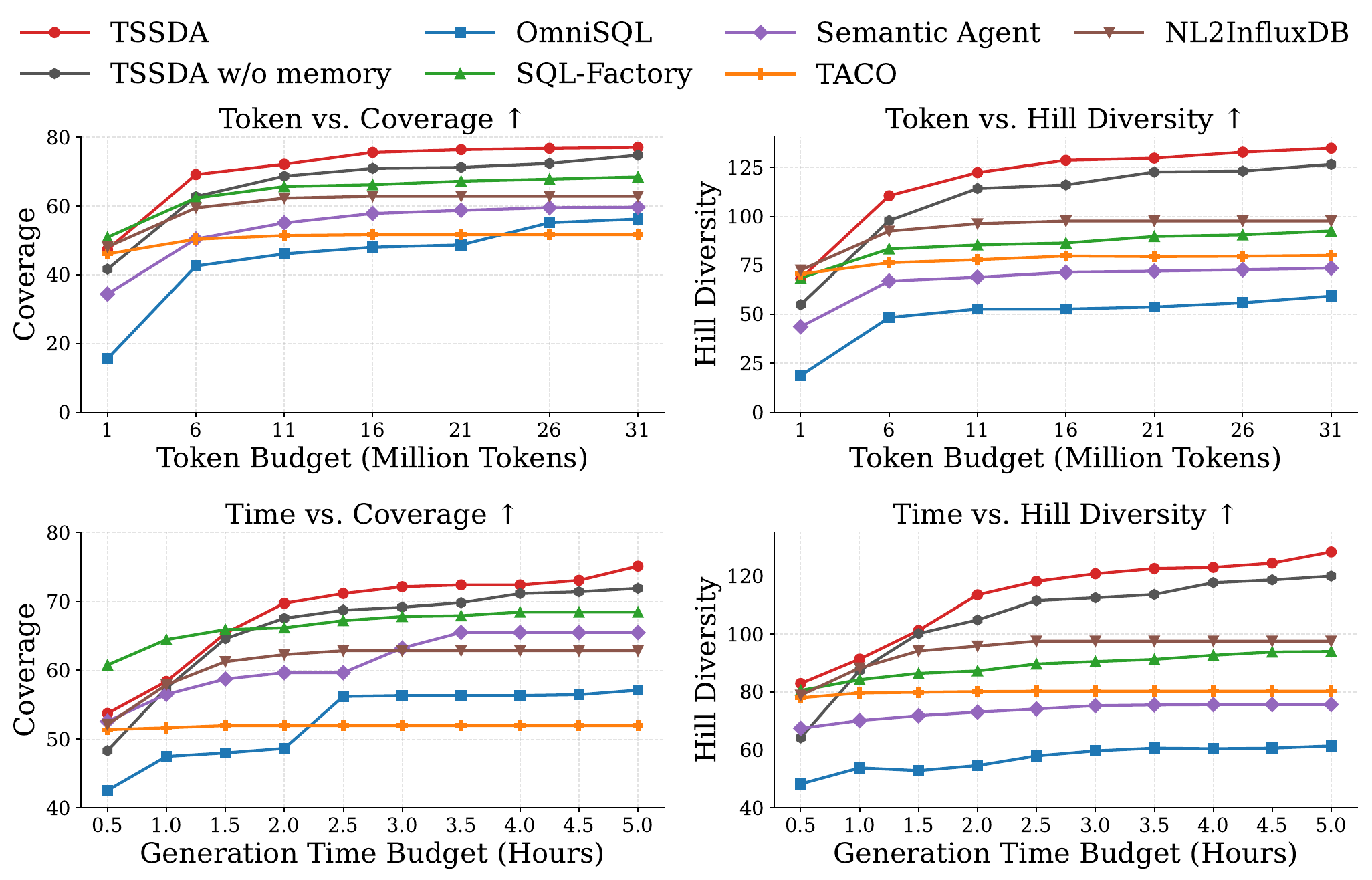}
  \caption{Efficiency Comparison of Different Methods}
  \label{fig:efficiency}
\end{figure}

To evaluate generation efficiency, we compare the TSSQL Coverage and Hill Diversity under varying resource budgets. We separately vary the token budget from 1 to 31 million tokens and the generation time budget from 0.5 to 5 hours. Figure~\ref{fig:efficiency} presents the results: 

\noindent (1) As token and time budgets increase, \modelName{} consistently outperforms all baselines on both metrics from 6M tokens and 2 hours onward, indicating more effective use of generation resources to achieve broader feature coverage and greater dataset diversity.

\noindent (2) \modelName{} achieves comparable effectiveness with substantially fewer resources. At 6M tokens or 2 hours, it surpasses all baselines at their respective budgets of 31M tokens or 5 hours.

\noindent (3) All methods generally improve with increasing token and time budgets, but most baselines, such as TACO and NL2InfluxDB, plateau early. In contrast, \modelName{} continues to improve as resources increase, widening its lead over these methods and indicating more efficient scaling of data generation.

\noindent (4) The \textit{w/o memory} variant shows lower TSSQL Coverage and Hill Diversity under matched budgets, supporting the efficiency benefits of the memory mechanism. Nevertheless, it still outperforms all baselines on both metrics at larger budgets.


\section{Related Work}

\subsection{Data Generation for NL2SQL}

Data generation for NL2SQL can be broadly divided into two categories: NL2SQL benchmark construction and SQL generation. We discuss these two categories in detail below.

\noindent \textbf{NL2SQL benchmark construction: }Early NL2SQL benchmarks~\cite{Spider, BIRD, SParC, KaggleDBQA} relied mainly on manual collection and annotation. However, manual construction requires substantial human effort and becomes particularly inefficient in the era of LLMs. To address this limitation, prior work has explored automated benchmark generation through predefined templates or data augmentation. DBPal~\cite{DBPal} generates NL--SQL pairs by instantiating templates with schema elements, while HIER~\cite{HIER} synthesizes SQL from abstract syntax trees and generates questions using a neural model. With the rise of LLMs, recent work~\cite{BenchPress, AMBROSIA, VLD-Bench, SQLMorph, PAE} has increasingly leveraged LLMs to assist benchmark construction. CLARITY~\cite{CLARITY} uses a schema-constrained LLM to generate ambiguous queries and multi-turn conversations from Spider and BIRD seeds. SQL-Exchange~\cite{SQL-Exchange} employs template-guided query transformations to generate samples through cross-database transfer. However, these approaches still rely on seed data or templates, with LLMs primarily serving as auxiliary tools. Moreover, none of the above approaches target TSDBs. Accordingly, a few studies have begun to investigate generating time-series queries from natural language for TSDBs. TEFD~\cite{TEFD} synthesizes Flux~\cite{influxDB} queries from manually predefined query templates. However, Flux is a pipeline-oriented query language rather than an SQL dialect. NL2InfluxDB~\cite{TS-NL2SQL} provides InfluxDB~\cite{influxDB} documentation to an LLM and directly generates queries. However, it produces only a small dataset through direct LLM synthesis, limiting systematic evaluation.

\noindent \textbf{SQL generation: }Beyond NL2SQL benchmark construction, SQL generation also involves automated query synthesis, but primarily aims to generate large-scale SQL workloads. Early SQL generation methods construct workloads by instantiating predefined templates or grammars~\cite{RAGS,QGEN,TPC-DS}. TPC-H~\cite{TPC-H} samples parameters for expert-designed query templates over a fixed schema. JOB~\cite{JOB} grounds workloads in real-world data using 33 query structures with predicate variants, while DSB~\cite{DSB} extends TPC-DS~\cite{TPC-DS} with configurable workload distributions to guide parameter instantiation. Despite increasing workload realism and control, these methods remain constrained by predefined query structures. Recent methods~\cite{Persona,SENSE,SQLStorm,SQLBarber} use LLMs to enable more flexible and scalable query synthesis. OmniSQL~\cite{OmniSQL} builds databases from web tables and synthesizes 2.54M pretraining samples through LLM-based CoT. SQL-Factory~\cite{SQL-Factory} introduces a multi-agent framework for data generation, expansion, rewriting, and validation. However, although these approaches introduce LLM-based agents, their autonomy remains largely confined to predefined pipelines and procedures. Consequently, they cannot autonomously control the generation process according to the evolving data environment, limiting data coverage and diversity. In contrast, we model the entire generation process as an interactive and explorable data space, transforming the agent from a passive data consumer into a Data Agent capable of exploring, orchestrating, and managing data generation.

\subsection{NL2SQL Methods}

Early NL2SQL methods~\cite{NaLIR,ATHENA} mapped natural language to SQL using hand-crafted semantic representations and domain knowledge. With the advancement of neural networks, subsequent studies~\cite{CatSQL, RAT-SQL, PICARD, PYTHIA} began to predict SQL using architectures such as LSTMs and attention mechanisms. However, limited model capabilities and generalization constrained their practical applicability. The emergence of LLMs substantially accelerated the development of NL2SQL~\cite{DAIL-SQL, Metasql}. DIN-SQL~\cite{DIN-SQL} employs in-context learning to decompose Text-to-SQL into multiple modules, such as schema linking and SQL generation. FINER-SQL~\cite{FINER-SQL} enhances SLMs through reasoning distillation and reinforcement learning with fine-grained execution feedback. More recent studies~\cite{SQLens, CHASE-SQL, DexterSQL} have increasingly focused on multi-stage reasoning. OpenSearch-SQL~\cite{OpenSearch-SQL} introduces a four-stage workflow with an alignment mechanism to maintain pipeline consistency. DeepEye-SQL~\cite{DeepEye-SQL} follows the software development lifecycle, integrating predefined reasoning, deterministic verification, and execution-aware selection. However, these methods focus on schema grounding and reasoning in relational databases, leaving TSSQL syntax and semantics, temporal operators, and specialized TSDB schemas insufficiently addressed.


\section{Conclusion}

We present \textbf{\modelName{}}, a Data Agent framework for Text-to-TSSQL data generation, and construct \textbf{\BenchName{}}, a large-scale benchmark. First, we systematically model TSSQL syntax and semantics to construct a data generation space, ensuring broad coverage and validity. Next, we develop a Data Agent that autonomously explores and orchestrates data generation through continuous decision-making, query generation, and adaptation, improving data diversity and complexity. Finally, we design a temporal-aware memory mechanism that improves generation efficiency by reusing historical experience. Extensive experiments on real-world industrial datasets demonstrate that Text-to-TSSQL remains challenging, while \textbf{\modelName{}} shows the best performance among the baseline methods.



\clearpage

\bibliographystyle{ACM-Reference-Format}
\bibliography{ref}

\end{document}